\documentclass[sigconf]{aamas} 

\usepackage{balance}
\usepackage{booktabs}
\usepackage{graphicx}
\usepackage{xcolor}
\usepackage{algorithm}
\usepackage{algorithmic}
\usepackage{amsmath}
\usepackage{mdframed}

\newmdenv[
  backgroundcolor=green!8,
  linecolor=green!40!black,
  linewidth=0.5pt,
  roundcorner=5pt,
  innerleftmargin=8pt,
  innerrightmargin=8pt,
  innertopmargin=6pt,
  innerbottommargin=6pt
]{insightbox}

\setcopyright{none}
\acmConference[ALA '26]{Submitted to the Adaptive and Learning Agents Workshop (ALA 2026) at AAMAS 2026}{May 25 -- 26, 2026}
{Paphos, Cyprus}{}
\copyrightyear{2026}
\acmYear{2026}
\acmDOI{}
\acmPrice{}
\acmISBN{}

\title[MAPLE: Decomposing Agent Adaptation]{MAPLE: A Sub-Agent Architecture for Memory, Learning, and Personalization in Agentic AI Systems}

\author{Deepak Babu Piskala}
\affiliation{
  \city{Seattle, WA}
  \country{USA}}

\begin{abstract}
Large language model (LLM) agents have emerged as powerful tools for complex tasks, yet their ability to adapt to individual users remains fundamentally limited. We argue this limitation stems from a critical architectural conflation: current systems treat memory, learning, and personalization as a unified capability rather than three distinct mechanisms requiring different infrastructure, operating on different timescales, and benefiting from independent optimization. We propose MAPLE (Memory-Adaptive Personalized LEarning)\footnote{Code: \url{https://github.com/prdeepakbabu/maple-framework}}, a principled decomposition where Memory handles storage and retrieval infrastructure; Learning extracts intelligence from accumulated interactions asynchronously; and Personalization applies learned knowledge in real-time within finite context budgets. Each component operates as a dedicated sub-agent with specialized tooling and well-defined interfaces. We ground our architecture in recent advances including neurobiologically-inspired memory~\cite{gutierrez2024hipporag}, cognitive agent frameworks~\cite{sumers2023cognitive}, and retrieval-augmented generation~\cite{lewis2020rag,gao2023retrieval}. Experimental evaluation on the MAPLE-Personas benchmark demonstrates that our decomposition achieves a 14.6\% improvement in personalization score compared to a stateless baseline ($p < 0.01$, Cohen's $d = 0.95$) and increases trait incorporation rate from 45\% to 75\%---enabling agents that genuinely learn and adapt.
\end{abstract}


\begin{document}

\pagestyle{fancy}
\fancyhead{}

\maketitle 


\section{Introduction}

Two employees open their company's AI assistant on the same Monday morning. Sarah, a senior ML engineer with five years of experience building production systems, types: ``What is a transformer in AI?'' Across the building, Marcus, a product manager three weeks into his first AI role, asks the exact same question.

The assistant gives them identical responses---a high-level overview with basic analogies about attention mechanisms. Sarah sighs and closes the window; she wanted implementation details about multi-head attention for a deployment she's debugging. Marcus copies the response but knows he'll need a colleague to translate the technical terms he doesn't recognize. This scene plays out thousands of times daily across enterprise AI deployments.

Here's the frustrating part: Sarah had already expressed her preferences. Last week, she gave a thumbs-down to an oversimplified explanation of gradient descent. The system stored this feedback. Marcus had been consistently engaging more with responses that used analogies, asking follow-up questions whenever technical jargon appeared. The system had these signals too. It ignored both.

When we say the AI ``doesn't remember,'' we're actually pointing at three different failures. We want it to \emph{store} what it learned about us---that's memory. We want it to \emph{extract insights} from our interactions---that's learning. And we want it to \emph{behave differently} based on who's asking---that's personalization. Most practitioners conflate these into one thing called ``memory.'' That conflation is precisely why systems disappoint.

\subsection{The Three Capabilities}

Let us be precise using Sarah and Marcus as our running example.

\textbf{Memory} is infrastructure---the organizational component responsible for storage and retrieval. When Sarah indicated frustration with simplified responses, memory preserves that signal. Memory answers: ``What do we know about this user?'' But memory doesn't decide what to store or how to apply it.

\textbf{Learning} extracts intelligence. When the system notices Sarah consistently skips introductions and jumps to code, learning identifies that as a meaningful pattern. When Marcus's follow-ups keep asking for simpler explanations, learning concludes he prefers accessible language. Learning answers: ``What should we know based on what just happened?'' But learning doesn't apply insights in real-time.

\textbf{Personalization} applies what we've learned. When Sarah asks about transformers, personalization retrieves her preference for technical depth and modifies the response. When Marcus asks, personalization shapes a different answer. Personalization answers: ``Given what we know, how should we respond differently?''

These three form a closed loop, but they must remain \emph{architecturally distinct}. Memory without learning means repeating errors forever---the system stores logs but never improves~\cite{packer2023memgpt}. Learning without personalization means optimizing for an average user that doesn't exist, frustrating everyone who deviates from the norm~\cite{salemi2024lamp}. And personalization without memory is simply impossible---you cannot adapt to someone you do not know.

\begin{insightbox}
Memory stores what we know. Learning extracts what we should know. Personalization applies what we've learned. Each requires different infrastructure, operates on different timescales, and benefits from independent optimization.
\end{insightbox}

\subsection{A Running Example}

Throughout this paper, we trace how MAPLE handles Sarah and Marcus differently. When Sarah asks about transformers:

\begin{enumerate}
    \item \textbf{Memory} retrieves her stored profile: senior ML engineer, PyTorch expertise, debugging a deployment, previously expressed preference for code examples
    \item \textbf{Learning} (having processed her past sessions) has extracted: ``Sarah engages more with implementation details; simplified explanations correlate with negative feedback''
    \item \textbf{Personalization} assembles context emphasizing technical depth, then generates: ``Here's the PyTorch implementation of multi-head attention, with notes on the memory layout issues you'll hit in production...''
\end{enumerate}

For Marcus, the same architecture produces different behavior:
\begin{enumerate}
    \item \textbf{Memory} retrieves: product manager, new to AI, asked many follow-up questions about terminology
    \item \textbf{Learning} has extracted: ``Marcus responds well to analogies; technical jargon without definition correlates with confusion''
    \item \textbf{Personalization} generates: ``Think of a transformer like a team of readers, each focusing on different parts of a document simultaneously. The `attention' mechanism is how they decide what's most relevant...''
\end{enumerate}

Same query, architecturally different responses, because the system decomposes adaptation into distinct capabilities.

\begin{figure}[t]
\centering
\includegraphics[width=\columnwidth]{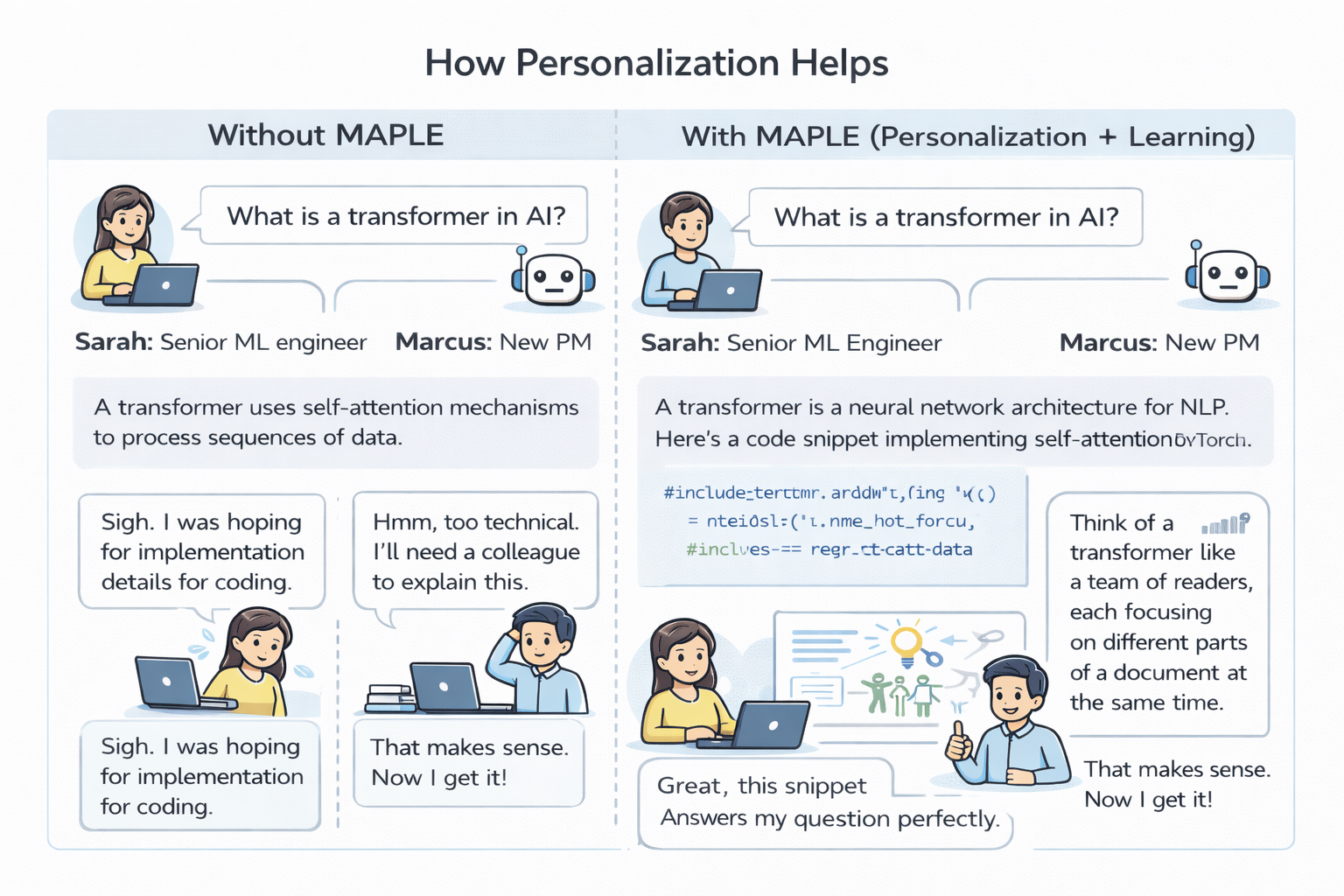}
\caption{Personalization in action: Sarah (senior ML engineer) and Marcus (product manager) ask the same question about transformers. MAPLE retrieves different user profiles and generates tailored responses---technical implementation details for Sarah, conceptual analogies for Marcus.}
\label{fig:sarah_marcus}
\end{figure}

Figure~\ref{fig:sarah_marcus} illustrates how the same query produces different responses for Sarah and Marcus. The MAPLE architecture enables this differentiation through its decomposed approach to memory, learning, and personalization.

\section{Background and Related Work}

\subsection{Memory Architectures for LLM Agents}

\textbf{MemGPT}~\cite{packer2023memgpt} introduces virtual context management inspired by operating system memory hierarchies. The key insight: LLMs face constraints analogous to limited RAM---the context window cannot hold all relevant information. MemGPT manages different memory tiers using the LLM itself to decide when to page information in and out. However, MemGPT focuses on the \emph{storage} dimension without distinguishing learning from personalization.

Consider how MemGPT would handle Sarah. It might page in her recent conversation history when context fills up, but it wouldn't extract that her negative feedback indicates a preference for technical depth. The tiered storage is sophisticated; the intelligence extraction is absent.

\textbf{Generative Agents}~\cite{park2023generative} present an architecture where agents maintain memory streams of all observations, synthesize memories into higher-level reflections, and retrieve relevant memories to plan behavior. Retrieval uses a scoring function combining recency, importance, and relevance. This work emphasizes \emph{reflection}---what we term learning---but integrates it tightly with memory storage without clear architectural separation.

\textbf{HippoRAG}~\cite{gutierrez2024hipporag} takes inspiration from hippocampal indexing theory of human long-term memory. The system orchestrates LLMs, knowledge graphs, and Personalized PageRank to mimic the different roles of neocortex (pattern storage) and hippocampus (index and retrieval). HippoRAG outperforms iterative retrieval methods while being 10-30x more efficient on multi-hop reasoning tasks.

\textbf{Recursive Summarization}~\cite{wang2023recursive} addresses long-term dialogue memory by recursively generating summaries: $m_{t+1} = \text{summarize}(m_t, c_{t+1})$. This progressive compression informs our approach to hierarchical memory.

\subsection{Cognitive Architectures}

\textbf{CoALA}~\cite{sumers2023cognitive} draws on cognitive science to propose a systematic framework for language agents with modular memory components: working memory (current context), episodic memory (specific events), semantic memory (general facts), and procedural memory (skills). This cognitive typology, grounded in Tulving's foundational work~\cite{tulving1972episodic}, provides valuable taxonomic grounding.

\textbf{ReAct}~\cite{yao2022react} establishes the Thought-Action-Observation pattern for agent reasoning. \textbf{Reflexion}~\cite{shinn2023reflexion} enables agents to reflect on past episodes and verbal feedback, improving performance through self-reflection rather than weight updates. Our Learning component similarly performs reflection on interaction histories---extracting patterns without gradient-based training. Recent surveys~\cite{wang2023survey,xi2023rise} identify memory as a key component of autonomous agents but treat it as monolithic. Multi-agent frameworks like AutoGen~\cite{wu2023autogen} and MetaGPT~\cite{hong2023metagpt} provide infrastructure for agent coordination but treat personalization as implicit.

\subsection{Personalization in Language Models}

LaMP~\cite{salemi2024lamp} benchmarks personalization across tasks, finding that retrieval-based personalization outperforms fine-tuning for many scenarios. PEARL~\cite{mysore2023pearl} uses generation-calibrated retrievers for writing assistance. User modeling has deep roots~\cite{rich1979user,fischer2001user} predating LLMs. MAPLE reconnects with this tradition: our user models make preferences explicit and editable rather than opaque.

\textbf{Memory Infrastructure.} Tools like Mem0 and LlamaIndex provide memory storage and retrieval infrastructure for LLM applications. However, these systems implement only the storage dimension---they do not address \emph{what} to learn from stored data (Learning) or \emph{how} to apply retrieved information to personalize responses (Personalization). MAPLE's decomposition is orthogonal to such infrastructure; our Memory sub-agent could be implemented using any of these storage backends. Prior work on agent architecture design~\cite{piskala2026files,piskala2025intelligence} argues for file-based interfaces and principled decomposition of intelligence---perspectives that inform our approach.

\begin{figure}[t]
\centering
\includegraphics[width=\columnwidth]{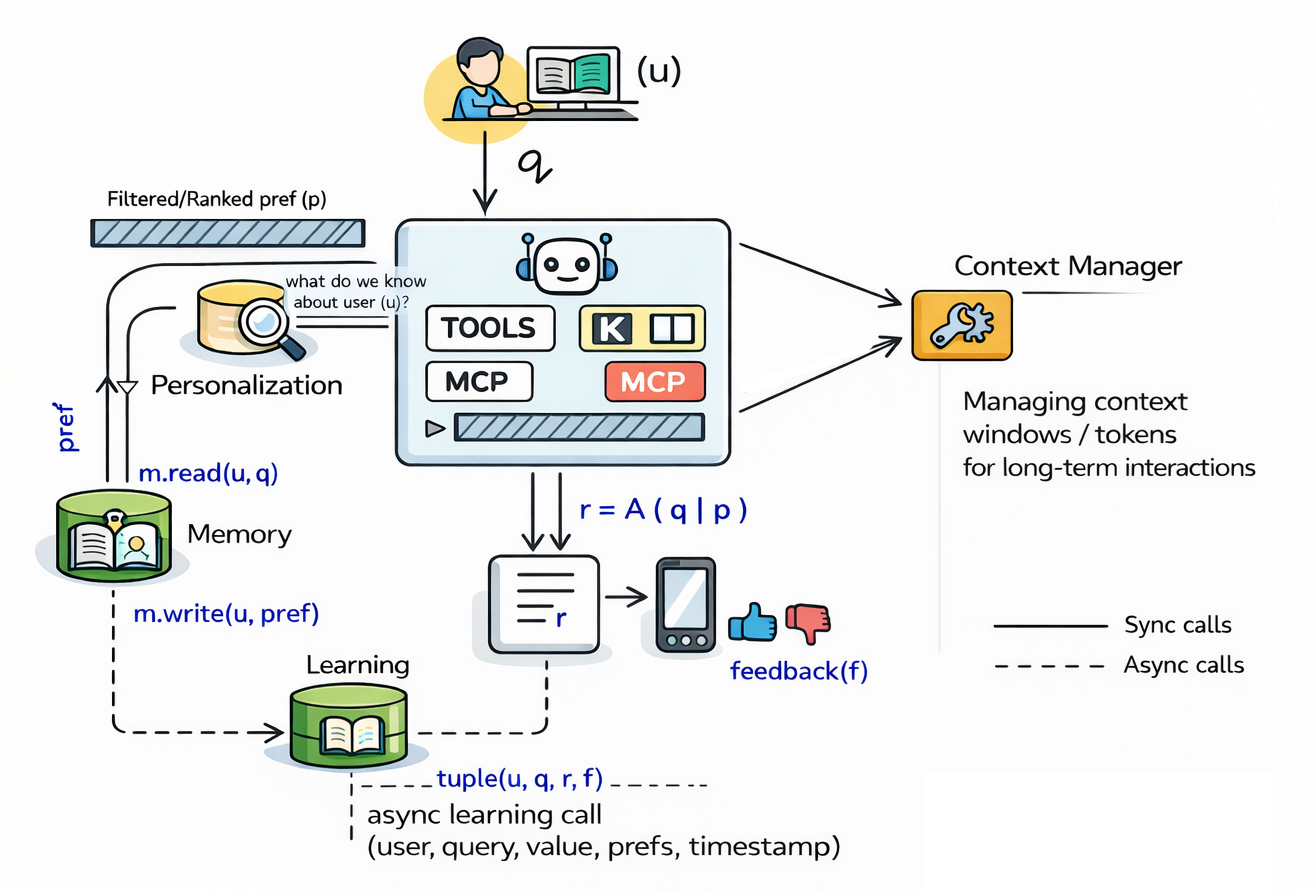}
\caption{The evolution from foundation models to personalized agents. Base LLMs provide general capabilities; RAG adds knowledge retrieval; MAPLE introduces memory, learning, and personalization as distinct architectural components, enabling agents that adapt to individual users over time.}
\label{fig:foundation}
\end{figure}

Figure~\ref{fig:foundation} positions our contribution within the landscape of foundation model capabilities. While base LLMs provide general intelligence and RAG adds external knowledge retrieval, MAPLE introduces the critical layer of user-specific adaptation---enabling agents that remember, learn from, and personalize to individual users.

\section{The MAPLE Framework}

\begin{figure*}[t]
\centering
\includegraphics[width=\textwidth]{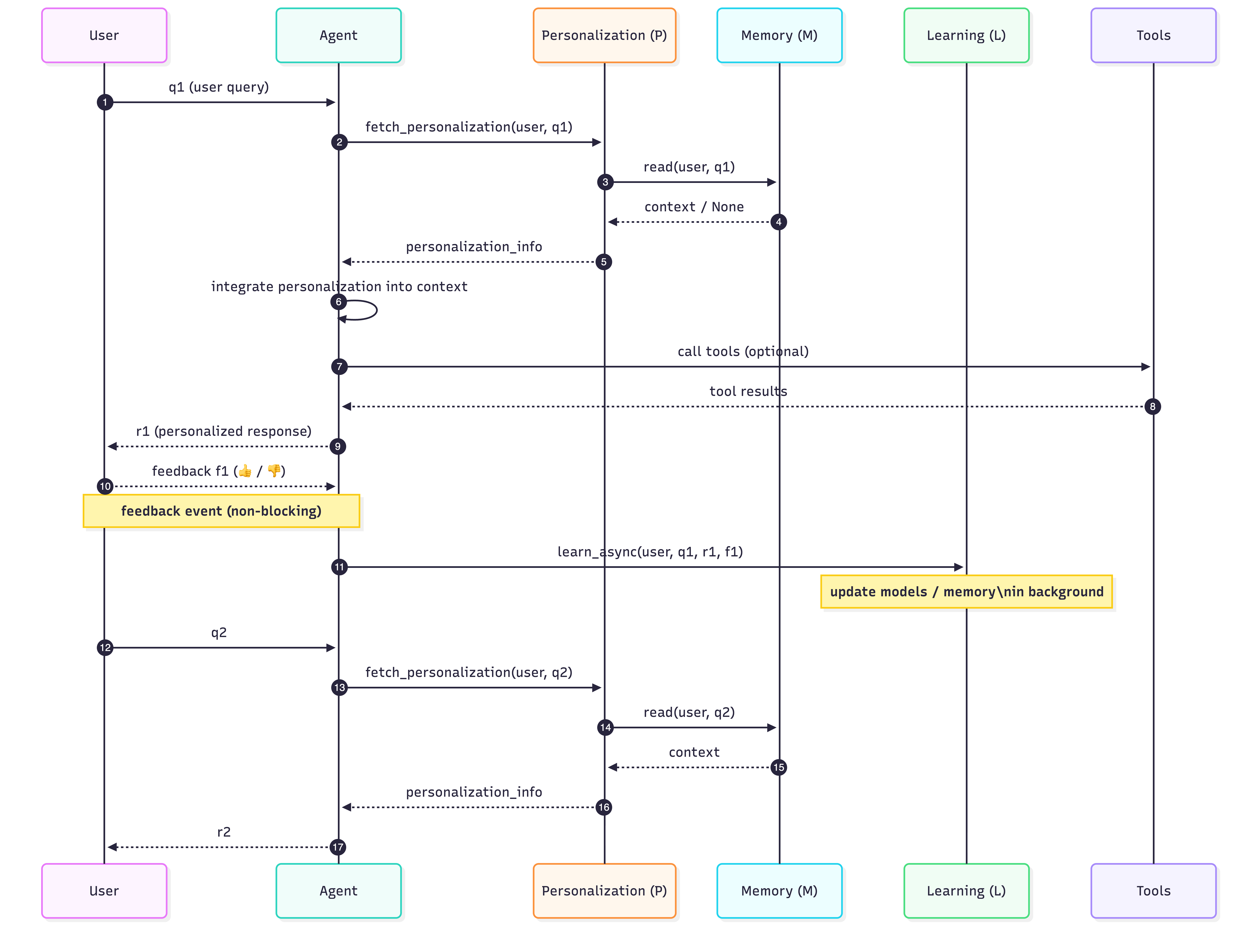}
\caption{MAPLE architecture as a sequence diagram showing the request-time flow (steps 1--4) and background learning loop (steps 5--6).}
\label{fig:architecture}
\end{figure*}

We now present our decomposition. The key insight: while Memory, Learning, and Personalization are functionally inter-dependent---forming a closed loop where memory enables learning enables personalization enables better memory---they require fundamentally different infrastructure. Figure~\ref{fig:architecture} illustrates the architecture as a sequence diagram. The flow proceeds as follows: (1) a user query arrives at the Agent; (2) the Agent consults Personalization; (3) Personalization retrieves relevant context from Memory; (4) the Agent generates a personalized response. Separately, a background learning loop operates asynchronously: (5) Learning analyzes completed sessions to extract insights; (6) insights are written back to Memory. This separation---real-time personalization in the request path, background learning outside it---is the key architectural distinction enabling both low latency and continuous improvement.

\subsection{Memory ($\mathcal{M}$): Storage Infrastructure}

Memory is the infrastructure of an agentic system---the organizational component responsible for storage and retrieval. If Learning determines \emph{what} to store and Personalization governs \emph{how to apply} what's stored, Memory handles the \emph{how} of storage itself---the data structures, indexing mechanisms, and retrieval protocols.

Think of Sarah's interaction when she expressed frustration with an oversimplified response. Three different systems engage: Learning extracts the insight (``Sarah prefers technical depth''), Personalization will eventually apply that insight to modify responses, but Memory is what \emph{preserves} the signal so it can be used later. Memory is the warehouse; Learning is the curator; Personalization is the shopper.

\subsubsection{Three Lenses for Classifying Memory}

Recent research proposes viewing memory through complementary taxonomies~\cite{sumers2023cognitive,tulving1972episodic}. Each lens reveals different architectural tradeoffs.

\textbf{Lens 1: By Form---What Carries the Memory?}

\emph{Token-level memory} ($\mathcal{M}_T$) consists of explicit, discrete units---text chunks, structured records, graph nodes. Sarah's preference might be stored as: \texttt{"prefers code examples over prose explanations"}. Token-level memory offers \emph{transparency} (exactly what's stored is visible), \emph{editability} (Sarah can correct or delete), and \emph{debuggability} (check the retrieved context to explain behavior).

\emph{Parametric memory} ($\mathcal{M}_\theta$) encodes information in model weights or LoRA adapters. A fine-tuned adapter might ``know'' to generate technical depth for users with Sarah's profile. Parametric memory offers \emph{seamlessness}---knowledge feels native---but sacrifices transparency.

\emph{Latent memory} ($\mathcal{M}_Z$) comprises hidden states, KV caches, and continuous embeddings persisting across turns~\cite{munkhdalai2024infini}. Latent memory is \emph{efficient} but \emph{ephemeral}---typically lost when sessions end.

Most production systems combine forms. Sarah's explicit preferences ($\mathcal{M}_T$) coexist with a domain-adapted model ($\mathcal{M}_\theta$) and session context ($\mathcal{M}_Z$).

\textbf{Lens 2: By Storage Structure (Within Token-level Memory)}

\emph{Flat (1D)}: Memories as unordered collections---vector databases, experience pools. Sarah's preferences sit alongside Marcus's in a single searchable index. Simple but relationships are implicit.

\emph{Planar (2D)}: Single-layer structures with explicit relationships---knowledge graphs, trees. Sarah's ``prefers technical depth'' connects to ``ML engineer'' connects to ``five years experience.'' Relationships are queryable.

\emph{Hierarchical (3D)}: Multi-layer structures with vertical abstraction~\cite{gutierrez2024hipporag}. Raw interactions aggregate into session summaries, which aggregate into user models, which connect to cohort patterns (``senior engineers tend to prefer...'').

\textbf{Lens 3: By Cognitive Analogy}

Following human memory research~\cite{tulving1972episodic,squire2004memory}:

\emph{Working memory}: Current session context---what Sarah asked three turns ago.

\emph{Episodic memory}: Specific events with temporal anchoring. ``Sarah asked about transformers on Monday while debugging a deployment issue.''

\emph{Semantic memory}: Abstracted facts, decontextualized knowledge. ``Sarah is an ML engineer who prefers technical depth.''

\emph{Procedural memory}: Skills and learned behaviors---templates that work well for users like Sarah.

Critical insight: episodic captures \emph{what happened}; semantic captures \emph{what we've concluded}. Systems need both. Episodic provides material for Learning; semantic provides efficient input for Personalization.

Here's the uncomfortable truth: most deployed AI systems don't learn. They retrieve. They cache. They pattern-match. But they don't \emph{change behavior} based on accumulated experience~\cite{wang2024comprehensive,ke2023continual}.

Our assistant might store every interaction Sarah ever had. But if it never extracts patterns, never identifies what made interactions successful, never generalizes---it has recall without growth. It's an archive, not a learner.

Memory is passive storage. Learning is active intelligence extraction. When Sarah expresses frustration with a simplified response, memory stores that signal. Learning asks: what does this tell us about Sarah's preferences? How should future behavior change? Does this pattern generalize to other users with similar profiles?

\textbf{Symbolic vs. Gradient-Based Learning.} We explicitly distinguish our approach from gradient-based learning that updates model parameters. MAPLE's Learning component performs \emph{symbolic learning}---extracting and storing insights in external data structures (knowledge graphs, user profiles) rather than updating neural network weights. This design choice offers two advantages: (1) \emph{No catastrophic forgetting}---insights about Sarah don't overwrite knowledge about Marcus, since they reside in separate database entries rather than shared parameters; (2) \emph{Immediate editability}---users can inspect, correct, or delete learned preferences, enabling transparency and control that parametric approaches cannot provide. Following Reflexion~\cite{shinn2023reflexion}, our Learning component reflects on interaction histories through LLM-based analysis, storing conclusions as structured facts rather than weight updates.

\subsubsection{What Learning Extracts}

\textbf{Facts} are objective truths. Learning infers that Sarah works on the ML platform team, has five years of experience, and is based in Seattle. Some facts come explicitly---Sarah states ``I'm a senior ML engineer.'' Others require inference---detecting PyTorch familiarity from code-heavy questions.

The challenge: distinguishing \emph{signal} from \emph{noise}. Sarah mentioning she's ``grabbing coffee'' doesn't warrant storage; Sarah mentioning she's ``switching from TensorFlow to JAX'' does.

\textbf{Preferences} capture how users like to interact. Learning extracts that Sarah prefers code examples while Marcus prefers analogies. These emerge from:
\begin{itemize}
    \item \emph{Explicit signals}: Sarah says ``I prefer shorter responses'' or clicks ``technical'' in settings
    \item \emph{Implicit signals}: Sarah consistently skips introductions $\Rightarrow$ prefers directness; Marcus's follow-ups always ask for clarification when jargon appears $\Rightarrow$ prefers accessible language
\end{itemize}

Implicit learning captures what users wouldn't think to state but carries risks: the system might misinterpret (Sarah skips intros because she's time-pressed, not because she finds them condescending).

\textbf{Experiences} are specific interactions worth preserving. Sarah's transformer question worked with a code-heavy response---that episode becomes relevant context for similar future questions.

\subsubsection{Levels of Learning}

\textbf{Level 1---Replay}: Store successful trajectories, retrieve by similarity. Sarah's transformer question worked; imitate that pattern for attention mechanisms. This is case-based reasoning~\cite{wang2023survey}---powerful when situations match, brittle otherwise.

\textbf{Level 2---Strategy Extraction}: Identify patterns across cases. ``Users asking about architecture while debugging prefer responses connecting theory to their specific errors.'' Requires metacognition---examining cases to extract generalizable principles.

\textbf{Level 3---Skill Synthesis}: Create new capabilities~\cite{qin2023tool,wang2024executable}. If the assistant repeatedly generates similar code transformations for Sarah, it might synthesize a utility function---expanding its action space through accumulated experience.

\subsubsection{Asynchronous Operation}

Learning cannot happen in the real-time request path---it's too slow. Instead, it operates asynchronously across three timescales. At \emph{end-of-session}, when Sarah closes chat, Learning processes the complete session to extract insights. \emph{Periodic} batch jobs identify cross-user patterns, discovering that senior engineers generally prefer technical depth. \emph{Event-triggered} processing handles critical explicit feedback (e.g., thumbs down) that bypasses queues for immediate updates to avoid repeating egregious mistakes.

The key architectural insight: Personalization and Memory operate in the request path (milliseconds), while Learning operates in the background (minutes to hours).

\subsection{Personalization ($\mathcal{P}$): Real-Time Adaptation}

Personalization is where memory and learning meet the individual user. It transforms ``an AI assistant'' into ``\emph{your} AI assistant.''

\subsubsection{Levels of Personalization}

\textbf{Surface personalization}: Names and basic preferences. ``Good morning, Sarah!'' Using preferred communication style. This is memory without learning---storing and retrieving explicit facts. Valuable for rapport, but shallow in impact.

\textbf{Behavioral personalization}: Adapting interaction patterns based on accumulated experience. The assistant learns that Sarah prefers code-first explanations while Marcus prefers conceptual overviews. For Sarah: lead with code examples, use technical terminology without definition, assume ML concept familiarity. For Marcus: start with analogies to business concepts, define technical terms on first use, build from basics to complexity.

\textbf{Goal personalization}: Understanding not just \emph{how} Sarah prefers information, but \emph{why} she's asking. Is she debugging? Learning? Preparing a presentation? The underlying goal shapes what information is relevant. This borders on theory of mind---modeling what users want, believe, are trying to accomplish.

\subsubsection{User Model Structure}

Effective personalization requires maintaining an explicit user model~\cite{fischer2001user}:
\begin{equation}
\mathcal{U}_u = (\mathcal{S}_u, \mathcal{D}_u, \mathcal{B}_u, \mathcal{G}_u)
\end{equation}

\begin{itemize}
    \item $\mathcal{S}_u$: Static attributes (role, team, tenure)---things that rarely change
    \item $\mathcal{D}_u$: Dynamic state (current goals, recent context, emotional tone)---Sarah was frustrated last session
    \item $\mathcal{B}_u$: Behavioral patterns (interaction preferences, engagement signals)---Sarah asks short questions expecting detailed answers
    \item $\mathcal{G}_u$: Predictive elements (anticipated needs)---Sarah's been asking about deployment; she'll probably want monitoring info soon
\end{itemize}

\subsubsection{The Personalization Mechanism}

When Sarah submits her query about transformers, the system follows a specific sequence:

\begin{enumerate}
    \item \textbf{Context analysis}: Analyze the current query---not just words, but signals about what Sarah might need
    \item \textbf{Selective retrieval}: Query Memory for preferences relevant to \emph{this specific question}. Her preference for technical depth matters; her preferred meeting times don't
    \item \textbf{Context assembly}: Load retrieved preferences into the LLM's context window as structured context: ``User preferences: prefers code over prose, expects implementation-level detail, has PyTorch expertise''
    \item \textbf{Instruction composition}: System prompt instructs the LLM to apply preferences: ``Adapt your response style based on user preferences. Lead with code examples and assume familiarity with ML concepts''
    \item \textbf{Generation}: LLM generates a response reflecting Sarah's preferences
\end{enumerate}

The same query from Marcus triggers different retrieval, loads different preferences, and produces a different response.

\subsection{The Closed Loop}

The system forms a closed feedback loop:
\begin{equation}
\mathcal{M} \xrightarrow{\text{retrieves}} \mathcal{P} \xrightarrow{\text{adapts}} r \xrightarrow{\text{feedback}} \mathcal{L} \xrightarrow{\text{writes}} \mathcal{M}
\end{equation}

Experience leads to reflection, reflection produces insight, insight drives behavior change, changed behavior generates new experience. Each interaction improves understanding. When Sarah submits her next query, Memory contains updated insights from previous learning. This is continual learning~\cite{wang2024comprehensive}---ongoing improvement without catastrophic forgetting.


\section{System Architecture}

We have constructed an agentic system that decomposes ``memory'' into three architecturally distinct sub-agents. Each operates as an independent component with specialized responsibilities, dedicated tooling, and its own LLM instance.

\textbf{The Agent} itself functions as an orchestrator with problem-solving capabilities. It maintains access to domain-specific tools (APIs, calculators, code executors) and knowledge bases. The Agent's primary responsibility is solving the user's problem. The P, M, L sub-agents \emph{augment} this core capability by making behavior adaptive rather than generic.

\textbf{Memory (M)} serves as storage and retrieval infrastructure. It maintains persistent user data across sessions---facts, preferences, interaction history. Memory answers ``what do we know?'' but makes no decisions about what deserves storage or how to apply it. Its tools include database connectors, vector stores, and knowledge graph interfaces.

\textbf{Learning (L)} operates as the intelligence extraction engine. It analyzes interactions to derive actionable insights, identifying patterns in behavior, inferring unstated preferences, and extracting generalizable strategies. Learning answers ``what should we know?'' but does not apply insights in real-time. It employs analytics engines, pattern recognition algorithms, and feedback processors.

\textbf{Personalization (P)} functions as the real-time adaptation layer. It retrieves relevant context from Memory and shapes system behavior. Personalization answers ``how should we respond differently?'' It operates through context managers, user model interfaces, and preference retrievers.

\vspace{-2ex}
\subsection{Design Rationale}

The sub-agent architecture follows three principles. \emph{Separation of concerns}: Memory does not decide what to learn, Learning does not apply insights during interactions, and Personalization does not determine what gets stored---enabling modular development and failure isolation. \emph{Asynchronous operation}: Memory and Personalization operate in the request path (adding 15-70ms latency through deterministic database queries), while Learning operates in the background (minutes to hours) with no impact on user-facing latency. \emph{Explicit user models}: Structured representations where individual preferences connect to role-based patterns, enabling both personalized and generalized reasoning.

\textbf{Why sub-agents?} This decomposition provides practical benefits: \emph{modularity}---Memory can be optimized for retrieval latency without affecting learning algorithms; \emph{specialization}---Personalization might use a fast, smaller model while Learning uses a larger model for complex pattern recognition; \emph{failure isolation}---if Learning encounters errors during background processing, active sessions continue unaffected; and \emph{observability}---each component's performance (retrieval latency, pattern accuracy, personalization effectiveness) becomes a distinct, measurable metric.

\vspace{-2ex}
\subsection{Implementation}

Our reference implementation demonstrates that MAPLE's architectural benefits emerge from the decomposition itself rather than sophisticated storage infrastructure~\cite{piskala2026files}. The Memory sub-agent uses lightweight filesystem storage with JSON-serialized records organized into three directories: \texttt{users/} (profile data), \texttt{episodic/} (session histories), and \texttt{semantic/} (extracted insights). This deliberate simplicity---no vector databases, no embedding models---shows that personalization gains come from the M/L/P decomposition, not from retrieval optimization. Production deployments could substitute any storage backend (Mem0, LlamaIndex, PostgreSQL) without architectural changes.


\section{Context Engineering}

The finite context window is the central constraint~\cite{chen2023extending,ding2024longrope}. System prompts, user context, retrieved documents, conversation history, tool definitions, and reasoning space all compete for limited attention. This is where \emph{context engineering} diverges from prompt engineering: while prompt engineering focuses on \emph{how} to phrase instructions, context engineering designs the architecture that feeds the \emph{right information at the right time}---ensuring the model has the right textbook before it starts thinking.

Bigger windows don't solve fundamental problems---they enable new failure modes we collectively term \emph{context rot}: poisoning (incorrect information compounds errors), distraction (over-reliance on stale patterns), confusion (irrelevant retrievals crowd useful ones), and clash (contradictory signals without resolution)~\cite{liu2024ring}. If Sarah's preference was incorrectly recorded, every subsequent interaction reinforces that error. These failures cannot be fixed with better prompts; they require architecture that actively curates what enters context, resolves conflicts, and allocates space dynamically.

\begin{figure}[!htbp]
\centering
\includegraphics[width=\columnwidth]{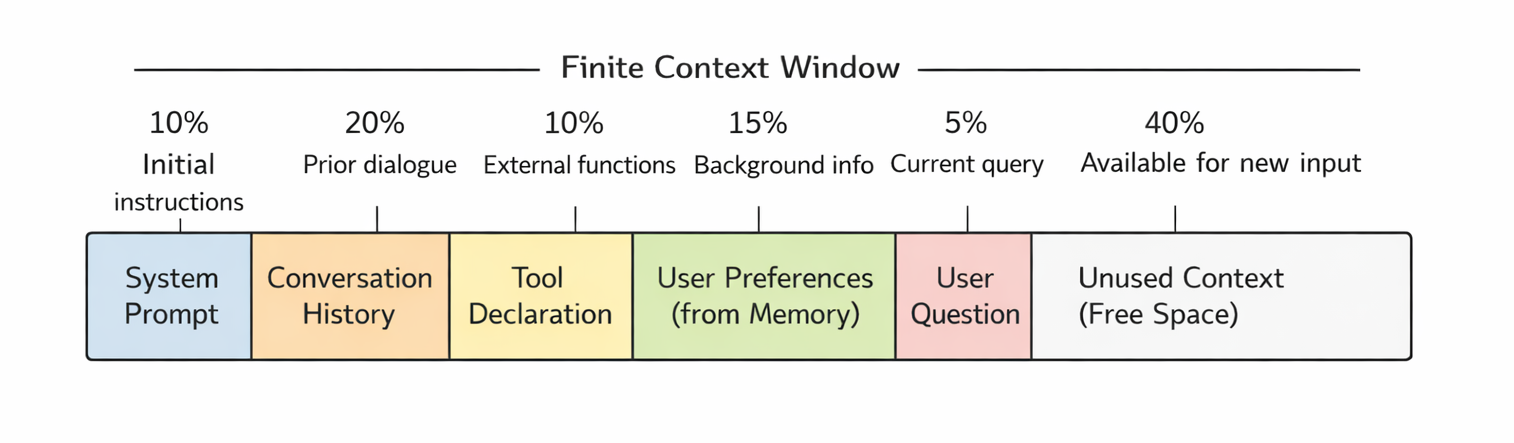}
\caption{Context window budget allocation. The finite window must accommodate system prompts (10\%), conversation history (20\%), tool declarations (10\%), user preferences from memory (15\%), the current query (5\%), and unused context for reasoning (40\%). Every token spent on personalization is unavailable for other components.}
\label{fig:context}
\end{figure}

MAPLE addresses context rot through intelligent context management. The Personalization component performs \emph{selective retrieval}---Sarah's technical depth preference matters for architecture questions; her meeting preferences don't. \emph{Hierarchical compression}~\cite{wang2023recursive} keeps recent interactions in detail while summarizing older ones. \emph{Conflict resolution} reasons about context when signals contradict, prioritizing recent explicit feedback over old implicit signals. Figure~\ref{fig:context} illustrates typical context window allocation where user preferences compete with other components within a finite budget.

\vspace{-2ex}
\section{Experimental Evaluation}

To empirically validate the MAPLE architecture, we conduct a controlled ablation study comparing systems with and without the Memory-Learning-Personalization decomposition. Our experimental design isolates the contribution of learned user context by evaluating response quality on held-out conversational turns where the system must proactively apply previously observed preferences.

\subsection{The MAPLE-Personas Benchmark}

We construct \textbf{MAPLE-Personas}, a novel benchmark designed to measure an agent's capacity for proactive personalization across multi-turn dialogues.\footnote{Dataset will be released at \url{huggingface.co/datasets/maple-personas}} The benchmark comprises 150 synthetically generated user personas, each characterized by a unique combination of five behavioral and demographic traits. These traits are sampled from a pool of 20 attributes spanning dietary preferences (e.g., vegetarian, health-conscious), living situations (e.g., has young children, cares for elderly parents), professional contexts (e.g., software engineer, works from home), environmental factors (e.g., cold climate, urban dwelling), and lifestyle characteristics (e.g., night owl, socially active, pet owner). The combinatorial sampling ensures diverse persona configurations that mirror the heterogeneity of real user populations.

For each of the 150 personas, we employ Claude 3 Haiku to synthesize a complete 10-turn conversational trajectory. The generation process is constrained to produce contextually coherent user utterances that organically reveal persona traits without explicit enumeration. Turns 1 through 8 constitute the \emph{learning phase}, during which user messages naturally expose their characteristics through domain-appropriate queries. For instance, a persona with the ``vegetarian'' trait might inquire about restaurant recommendations, while one with ``young children'' might ask about weekend activities. Turns 9 and 10 form the \emph{evaluation phase}, comprising test queries deliberately unrelated to previously discussed topics. This design creates a rigorous evaluation criterion: the system must incorporate knowledge from the learning phase into responses for novel queries, demonstrating genuine preference transfer rather than mere conversational continuity.

\subsection{Experimental Protocol}

We evaluate two system configurations under identical inference conditions. The \textbf{Baseline} condition employs the same underlying LLM (Claude 3 Sonnet) without persistent memory or learned user context. Each response is generated solely from the current turn's query, simulating a stateless assistant that treats every interaction as independent. The \textbf{MAPLE} condition activates the complete Memory-Learning-Personalization architecture. During the learning phase (turns 1--8), the Learning sub-agent asynchronously extracts user preferences and stores them via the Memory sub-agent. During the evaluation phase (turns 9--10), the Personalization sub-agent retrieves relevant context and injects it into the generation prompt, enabling preference-aware responses.

Both conditions process the same conversational inputs and are evaluated on identical output criteria. The experimental isolation ensures that observed performance differences are attributable solely to the presence or absence of the M/L/P decomposition.

\subsection{Evaluation Methodology}

We adopt the LLM-as-judge paradigm~\cite{zheng2023judging,li2023alpacaeval} to assess personalization quality at scale. For each evaluation turn, an independent judge model (Claude 3 Haiku) receives three inputs: the complete list of user traits revealed during the learning phase, the current query, and the assistant's response. The judge produces a structured assessment comprising a personalization score on a 5-point ordinal scale and a trait-level consistency analysis.

The \emph{Personalization Score} captures the degree to which the response proactively incorporates learned user context. A score of 5 indicates that multiple relevant traits are naturally woven into the response without explicit user prompting. A score of 4 reflects appropriate trait references that enhance relevance. A score of 3 denotes a generic but helpful response that neither leverages nor contradicts known preferences. Scores of 2 and 1 indicate missed personalization opportunities and active contradiction of established preferences, respectively.

The \emph{Trait Consistency Analysis} provides a fine-grained decomposition of how each revealed trait manifests in the response. For each of the user's five traits, the judge classifies it as \emph{incorporated} (actively used to shape content), \emph{violated} (contradicted by the response), or \emph{neutral} (not relevant to the current query). This enables computation of a Trait Incorporation Rate: the proportion of query-relevant traits that are successfully reflected in the response.

\subsection{Quantitative Results}

\begin{figure}[!htbp]
\centering
\includegraphics[width=\columnwidth]{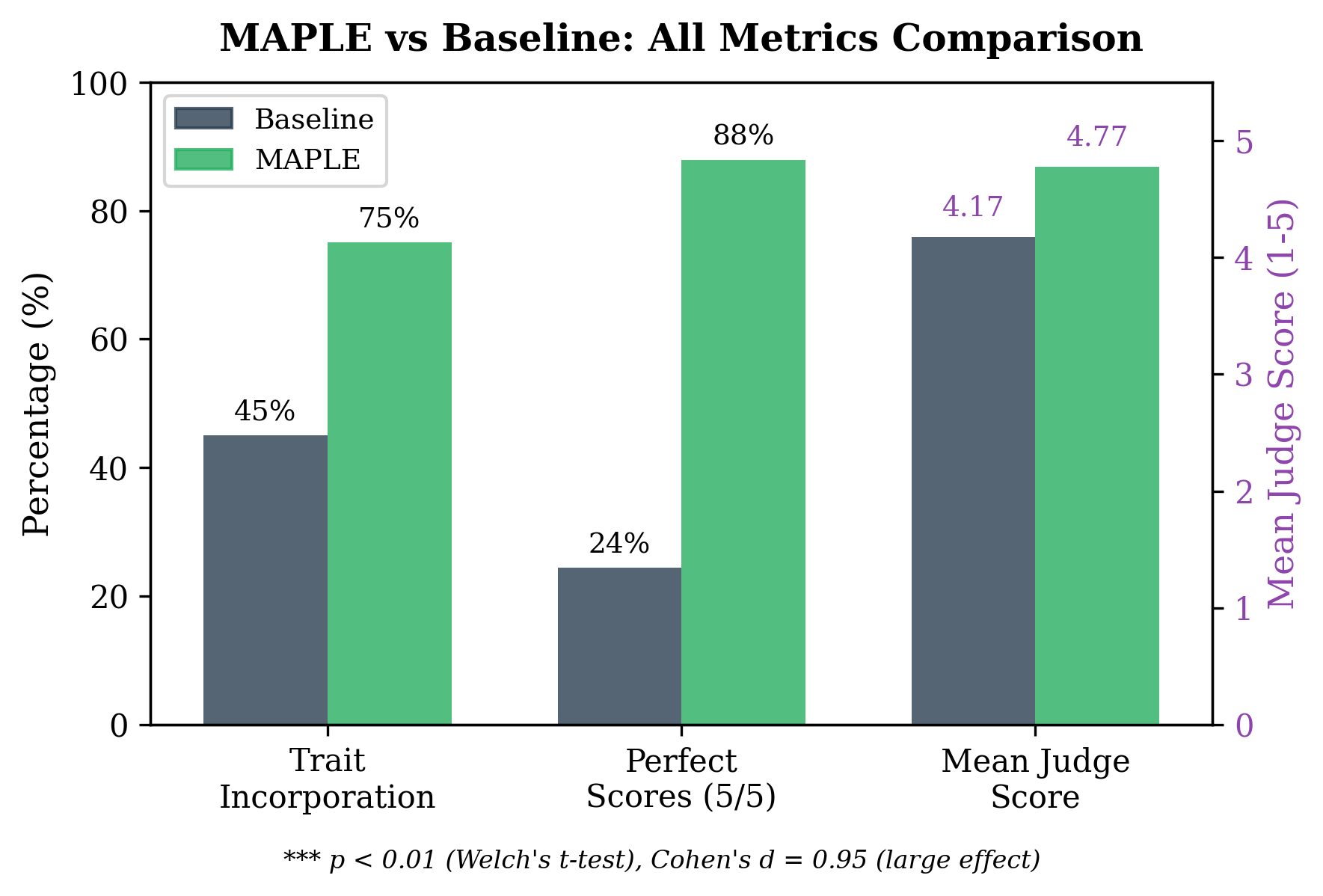}
\caption{MAPLE vs Baseline across all metrics. Left axis: percentage metrics (trait incorporation, perfect scores). Right axis: mean judge score (1--5 scale). MAPLE achieves statistically significant improvements across all measures (Cohen's $d = 0.95$, $p < 0.01$).}
\label{fig:results}
\end{figure}

Table~\ref{tab:results} summarizes the primary experimental outcomes. Figure~\ref{fig:results} visualizes the effect size and score distributions.

\begin{table}[!htbp]
\centering
\caption{Personalization quality comparison ($^{***}p < 0.01$).}
\label{tab:results}
\small
\begin{tabular}{lccc}
\toprule
\textbf{Metric} & \textbf{Base.} & \textbf{MAPLE} & \textbf{$\Delta$} \\
\midrule
Judge Score (1--5) & $4.17$ & $\mathbf{4.78}$ & \textcolor{green!60!black}{$+0.61^{***}$} \\
Trait Incorp. & 45\% & \textbf{75\%} & \textcolor{green!60!black}{+30pp} \\
Perfect (5/5) & 15\% & \textbf{88\%} & \textcolor{green!60!black}{+73pp} \\
\bottomrule
\end{tabular}
\end{table}

The results demonstrate a substantial and statistically significant advantage for MAPLE across all metrics. The mean personalization score improves from 4.17 to 4.78, representing a 14.6\% relative gain. We assess statistical significance using Welch's $t$-test for unequal variances, obtaining $t = 6.05$ with 74.5 degrees of freedom and $p < 0.001$. The effect size, measured by Cohen's $d = 0.95$, exceeds the conventional threshold for a large effect ($d \geq 0.8$), indicating that the performance difference is not merely statistically significant but practically meaningful.

The Trait Incorporation Rate rises from 45\% to 75\%, a 30 percentage point absolute improvement. Perhaps most strikingly, the proportion of responses receiving perfect personalization scores (5/5) increases from 15\% to 88\%, suggesting that MAPLE enables consistently high-quality personalized responses rather than occasional successes.

\subsection{Qualitative Analysis}

Beyond aggregate metrics, qualitative examination of individual responses reveals systematic behavioral differences between conditions. We present two representative examples that illustrate the nature of MAPLE's personalization advantage.

\textbf{Example 1: Multi-constraint Synthesis.} Consider a persona characterized by ``has two young children,'' ``lives in a big city,'' and ``has a dog named Max.'' When asked for living room redecorating suggestions---a query unrelated to any topic discussed during the learning phase---MAPLE responds: ``Based on your family situation with young children, a dog, and urban living, I recommend durable stain-resistant fabrics, pet-friendly materials that resist scratching, and space-efficient storage solutions for city apartments.'' The baseline, lacking this context, provides generic trend-based advice: ``Based on 2026 interior design trends, consider warm earth tones and statement lighting.'' MAPLE synthesizes three independent traits into a coherent, practically relevant response.

\textbf{Example 2: Implicit Need Anticipation.} A persona with traits ``lives alone,'' ``cold climate,'' and ``has a bad back'' receives the same redecorating query. MAPLE's response addresses implicit needs: ``I recommend ergonomic seating options for back support, warm cozy color palettes suited to cold climates, and comfortable single-person arrangements.'' The baseline offers sophisticated color schemes without considering the user's physical constraints or environmental context. This example demonstrates MAPLE's capacity to translate learned facts into actionable recommendations that anticipate unstated needs.

Across all qualitative examples, we observe consistent patterns distinguishing MAPLE from baseline responses. MAPLE responses characteristically open with explicit trait acknowledgment (``Based on your...''), whereas baseline responses reference external sources (``Based on trends...'') or request information the system should already possess. MAPLE incorporates 2--4 traits per response on average, while baseline responses incorporate zero.


\section{Discussion}

MAPLE embodies multi-agent design principles~\cite{wu2023autogen,hong2023metagpt,li2023camel}: role specialization, coordination through interfaces, asynchronous collaboration, and emergent adaptation.

\textbf{Limitations}: Our evaluation relies on synthetic personas and LLM-as-judge methodology. While this approach enables controlled experimentation at scale, it introduces potential self-preference bias when Claude-family models generate responses that Claude judges. We mitigate this by using different model variants (Sonnet for generation, Haiku for judging) and focusing on structured trait-level assessments rather than open-ended quality ratings. Future work should validate these findings with human evaluation and diverse judge models. Additionally, three sub-agents require coordination infrastructure, personalization quality remains subjective, and cross-user learning requires careful anonymization.

\textbf{Future Work}: Human evaluation studies to validate LLM judge assessments. Cross-model evaluation with diverse judge models (GPT-4, Llama) to test generalization. Federated learning for cross-user patterns with differential privacy. Longitudinal deployment studies in production environments.


\section{Conclusion}

We presented MAPLE, decomposing agent adaptation into Memory (storage infrastructure), Learning (intelligence extraction), and Personalization (real-time application). Through Sarah and Marcus, we demonstrated how decomposition enables differentiated responses from the same architecture.

The key insight: treating M, L, P as distinct subsystems with different computational characteristics and timescales enables principled development of agents that genuinely learn and adapt. MAPLE addresses the fundamental question for adaptive agents: not just what to remember, but what to learn from memories, and how to apply learning in the moment.



\bibliographystyle{ACM-Reference-Format} 
\bibliography{references}


\clearpage
\appendix

\section{Prompts and Implementation Details}

This appendix provides the prompts used by the Learning and Personalization components, along with illustrative examples of their operation.

\subsection{Learning Component Prompt}

The Learning component uses the following prompt template to extract insights from each conversation turn:

\begin{small}
\begin{verbatim}
Analyze this conversation turn and extract 
insights about the user.

User message: {user_message}
Assistant response: {assistant_response}
User feedback: {feedback}

Extract ANY useful insights about the user:
- PREFERENCE: How they want information 
  presented (with citations, concise, 
  detailed, code examples, etc.)
- FACT: Information about them (role, 
  expertise, company, domain interests)
- BEHAVIOR: Communication patterns (asks 
  follow-ups, prefers examples, etc.)

IMPORTANT: If the user explicitly states a 
preference or request about how they want 
information, ALWAYS capture it.

Return insights as JSON array:
[
  {"type": "preference|fact|behavior", 
   "content": "detailed insight text", 
   "confidence": 0.0-1.0}
]

Example - If user says "I want URLs for 
sources", return:
[{"type": "preference", 
  "content": "User wants citation URLs and 
  source links when facts are presented", 
  "confidence": 0.95}]

Return [] only if the message is a simple 
greeting with no information.
\end{verbatim}
\end{small}

\subsection{Personalization Context Structure}

The Personalization component assembles user context into a structured format injected into the LLM's system prompt. The template below shows placeholder variables (in curly braces) for profile fields, with an example instantiation for Sarah's context in the Facts, Preferences, and Summary sections:

\begin{small}
\begin{verbatim}
## User Context

### User Profile
- Role: {role}
- Expertise: {expertise_level}
- Preferred language: {language}
- Preferred style: {response_style}
- Appreciates code examples
- Verbosity preference: {verbosity}

### What You Know About This User
Use these facts and preferences when 
responding:

**Facts:**
- User is a senior ML engineer
- User works with PyTorch daily
- User is debugging a production deployment

**Preferences:**
- User prefers code examples over prose
- User wants technical depth without 
  introductory explanations
- User appreciates implementation details

**Communication patterns:**
- User asks short questions expecting 
  detailed answers
- User engages more with responses 
  containing runnable code

### Recent Conversation Summary
- User asked about: transformer architecture
- User asked about: attention mechanisms
- User asked about: deployment optimization
\end{verbatim}
\end{small}

\subsection{Illustrative Examples}

\subsubsection{Example 1: Learning Extraction}

\textbf{User message:} ``I'm a data scientist at a biotech company. Can you explain how transformers work? I'd prefer code examples in Python.''

\textbf{Learning output:}
\begin{small}
\begin{verbatim}
[
  {"type": "fact", 
   "content": "User is a data scientist", 
   "confidence": 0.95},
  {"type": "fact", 
   "content": "User works at a biotech 
   company", 
   "confidence": 0.95},
  {"type": "preference", 
   "content": "User prefers code examples 
   in Python", 
   "confidence": 0.90}
]
\end{verbatim}
\end{small}

\subsubsection{Example 2: Personalization in Action}

For Sarah (senior ML engineer), the Personalization component retrieves:
\begin{itemize}
    \item Facts: Senior ML engineer, PyTorch expertise, 5 years experience
    \item Preferences: Prefers code over prose, expects implementation-level detail
    \item Behaviors: Gives negative feedback to oversimplified responses
\end{itemize}

The assembled context instructs the LLM:
\begin{small}
\begin{verbatim}
Adapt your response to this user's 
preferences. Lead with code examples, use 
technical terminology without definition, 
assume familiarity with ML concepts. Avoid 
introductory explanations.
\end{verbatim}
\end{small}

For Marcus (product manager), different context is retrieved:
\begin{itemize}
    \item Facts: Product manager, new to AI, three weeks in role
    \item Preferences: Appreciates analogies, needs definitions for jargon
    \item Behaviors: Asks follow-up questions when technical terms appear
\end{itemize}

The assembled context instructs the LLM:
\begin{small}
\begin{verbatim}
Adapt your response to this user's 
preferences. Start with analogies and 
conceptual explanations. Define technical 
terms on first use. Build from basics to 
complexity.
\end{verbatim}
\end{small}

\subsubsection{Example 3: End-to-End Flow}

\textbf{Query:} ``What is a transformer in AI?''

\textbf{Sarah's Response (MAPLE):}
\begin{quote}
\small
Here's the PyTorch implementation of multi-head attention:
\begin{verbatim}
class MultiHeadAttention(nn.Module):
    def __init__(self, d_model, n_heads):
        super().__init__()
        self.d_k = d_model // n_heads
        self.W_q = nn.Linear(d_model, d_model)
        ...
\end{verbatim}
Key implementation details for production: (1) Use \texttt{scaled\_dot\_product\_attention} for memory efficiency; (2) Watch for the $O(n^2)$ memory scaling with sequence length...
\end{quote}

\textbf{Marcus's Response (MAPLE):}
\begin{quote}
\small
Think of a transformer like a team of readers working on a document together. Each reader (called an ``attention head'') focuses on different aspects---one might track who is doing what, another might follow the timeline of events.

The key innovation is ``attention''---instead of reading word by word like earlier AI models, transformers can look at all words simultaneously and decide which ones are most relevant to understanding each part...
\end{quote}

\textbf{Baseline Response (identical for both):}
\begin{quote}
\small
A transformer is a type of neural network architecture introduced in the paper ``Attention Is All You Need'' (2017). It uses self-attention mechanisms to process sequential data. The key components include multi-head attention, feed-forward layers, and positional encodings...
\end{quote}


\end{document}